\documentclass[11pt,a4paper]{article}
\usepackage[hyperref]{emnlp-ijcnlp-2019}
\usepackage{times}
\usepackage{latexsym}

\usepackage{url}

\usepackage [utf8]{inputenc}
\usepackage{booktabs}
\usepackage{multirow}

\usepackage{graphicx}
\usepackage{makecell}  
\usepackage{subcaption}
\usepackage{amsmath}
\usepackage{listings}
\usepackage[colorinlistoftodos,textsize=small]{todonotes}

\aclfinalcopy 

\title{EASSE: {E}asier {A}utomatic {S}entence {S}implification {E}valuation}

\author{Fernando Alva-Manchego \\ University of Sheffield \\ {\tt f.alva@sheffield.ac.uk}
        \And  Louis Martin  \\ Facebook AI Research \\ Inria \\ {\tt louismartin@fb.com} 
        \AND Carolina Scarton \\ University of Sheffield \\ {\tt c.scarton@sheffield.ac.uk}
        \And Lucia Specia \\ Imperial College London \\ {\tt l.specia@imperial.ac.uk}}

\date{}

\begin{document}
\maketitle
\begin{abstract}
We introduce EASSE, a Python package aiming to facilitate and standardise automatic evaluation and comparison of Sentence Simplification (SS) systems.
EASSE provides a single access point to a broad range of evaluation resources: standard automatic metrics for assessing SS outputs (e.g.\ SARI), word-level accuracy scores for certain simplification transformations, reference-independent quality estimation features (e.g.\ compression ratio), and standard test data for SS evaluation (e.g.\ TurkCorpus).
Finally, EASSE generates easy-to-visualise reports on the various metrics and features above and on how a particular SS output fares against reference simplifications.
Through experiments, we show that these functionalities allow for better comparison and understanding of the performance of SS systems. 
\end{abstract}

\section{Introduction}
Sentence Simplification (SS) consists of modifying the content and structure of a sentence to improve its readability while retaining its original meaning. 
For automatic evaluation of a simplification output, it is common practice to use machine translation (MT) metrics (e.g.\ BLEU \cite{papineni-etal:2002:Bleu}), simplicity metrics (e.g.\ SARI \cite{xu-etal:2016}), and readability metrics (e.g.\ FKGL  \cite{kincaid-etal:1975:FK}). 

Most of these metrics are available in individual code repositories, with particular software requirements that sometimes differ even in programming language (e.g.\ corpus-level SARI is implemented in Java, whilst sentence-level SARI is available in both Java and Python). 
Other metrics (e.g.\ SAMSA~\citep{samsa:sulem-etal:2018}) suffer from insufficient  documentation or require executing multiple scripts with hard-coded paths, which prevents researchers from using them.

EASSE ({E}asier {A}utomatic {S}entence {S}implification {E}valuation) is a Python package that provides access to popular automatic metrics in SS evaluation and ready-to-use public datasets through a simple command-line interface.
With this tool, we make the following contributions: (1) we provide popular automatic metrics in a single software package, (2) we supplement these metrics with word-level transformation analysis and reference-less Quality Estimation (QE) features, (3) we provide straightforward access to commonly used evaluation datasets, and (4) we generate a comprehensive HTML report for quantitative and qualitative evaluation of a SS system.
We believe this package will facilitate evaluation and improve reproducibility of results in SS.
EASSE is available in \url{https://github.com/feralvam/easse}.
 

\section{Package Overview} \label{sec:package}

\subsection{Automatic Corpus-level Metrics}
\label{sec:metrics}
Although human judgements on grammaticality, meaning preservation and simplicity are considered the most reliable method for evaluating a SS system's output~\citep{stajner-etal:2016:QATS}, it is common practice to use automatic metrics. 
They are useful for either assessing systems at development stage, to compare different architectures, for model selection, or as part of a training policy.
EASSE implementation works as a wrapper for the most common evaluation metrics in SS:

\paragraph{BLEU} is a precision-oriented metric that relies on the proportion of n-gram matches between a system's output and reference(s). 
Previous work~\cite{xu-etal:2016} has shown that BLEU correlates fairly well with human judgements of grammaticality and meaning preservation. 
EASSE uses \textsc{SacreBleu}~\citep{sacrebleu:post:2018}\footnote{\url{https://github.com/mjpost/sacreBLEU}} to calculate BLEU.
This package was designed to standardise the process by which BLEU is calculated: it only expects a detokenised system's output and the name of a test set.
Furthermore, it ensures that the same pre-processing steps are used for the system output and reference sentences. 

\paragraph{SARI} measures how the simplicity of a sentence was improved based on the words added, deleted and kept by a system.
The metric compares the system’s output to multiple simplification references and the original sentence. 
SARI has shown positive correlation with human judgements of simplicity gain. 
We re-implement SARI's corpus-level version in Python (it was originally available in Java).  
In this version, for each operation ($ope \in \{add, del, keep\}$) and $n$-gram order, precision $p_{ope}(n)$, recall $r_{ope}(n)$ and F1 $f_{ope}(n)$ scores are calculated. These are then averaged over the $n$-gram order to get the overall operation F1 score $F_{ope}$:
    \begin{align*}
    f_{ope}(n) &= \frac{2 \times p_{ope}(n) \times r_{ope}(n)}{p_{ope}(n)+r_{ope}(n)}\\
    F_{ope} &= \frac{1}{k}\sum_{n=[1,..,k]}f_{ope}(n)
    \end{align*}

\noindent{Although \citet{xu-etal:2016} indicate that only precision should be considered for the deletion operation, we follow the Java implementation that uses F1 score for all operations in corpus-level SARI.}

\paragraph{SAMSA} measures structural simplicity (i.e.\ sentence splitting). 
This is in contrast to SARI, which is designed to evaluate simplifications involving paraphrasing. 
EASSE re-factors the original SAMSA implementation\footnote{\url{https://github.com/eliorsulem/SAMSA}} with some modifications: 
(1) an internal call to the TUPA parser~\citep{hershcovich-etal:2017:TUPA}, which generates the semantic annotations for each original sentence;
(2) a modified version of the monolingual word aligner ~\citep{sultan-etal:2014} that is compatible with Python 3, and uses Stanford CoreNLP~\citep{corenlp:manning-etal:2014}\footnote{\url{https://stanfordnlp.github.io/stanfordnlp/corenlp_client.html}} through their official Python interface; and (3) a single function call to get a SAMSA score instead of running a series of scripts. 

\paragraph{FKGL} Readability metrics, such as Flesch-Kincaid Grade Level~ (FKGL), are commonly reported as measures of simplicity. 
They however only rely on average sentence lengths and number of syllables per word, so short sentences would get good scores even if they are ungrammatical, or do not preserve meaning~\citep{wubben-etal:2012}.
Therefore, these scores should be interpreted with caution.
EASSE re-implements FKGL by porting publicly available scripts\footnote{\url{https://github.com/mmautner/readability}} to Python 3 and fixing some edge case inconsistencies (e.g. newlines incorrectly counted as words or bugs with memoization).

\subsection{Word-level Analysis and QE Features}
\paragraph{Word-level Transformation Analysis}
\label{sec:wordlevel_analysis}

\begin{figure*}[tb]
    \centering
    \includegraphics[width=1.33\columnwidth]{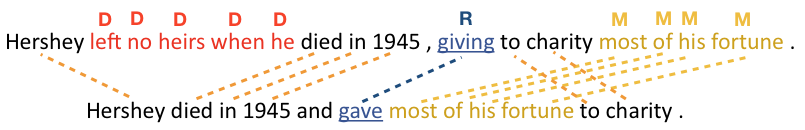}
    \caption{Example of automatic transformation annotations based on word alignments between an original (top) and a simplified (bottom) sentence. Unaligned words are \textsc{delete}. Words that are aligned to a different form are \textsc{replace}. Aligned words without an explicit label are \textsc{copy}. A word whose relative index in the original sentence changes in the simplified one is considered a \textsc{move}.\label{fig:annotation}}
\end{figure*}


EASSE includes algorithms to determine which specific text transformations a SS system performs more effectively. This is done based on word-level alignment and analysis.


Since there is no available simplification dataset with manual annotations of the transformations performed, we re-use the annotation algorithms from MASSAlign~\cite{paetzold-etal:2017:massalign}.
Given a pair of sentences (e.g. original and system output), the algorithms use word alignments to identify deletions, movements, replacements and copies (see Fig.~\ref{fig:annotation}).
This process is prone to some errors: when compared to manual labels produced by four annotators in 100 original-simplified pairs, the automatic algorithms achieved a micro-averaged F1 score of 0.61~\citep{alvamanchego:2017:ijcnlp}.

We generate two sets of automatic word-level annotations: (1) between the original sentences and their reference simplifications, and (2) between the original sentences and their automatic simplifications produced by a SS system.
Considering (1) as reference labels, we calculate the F1 score of each transformation in (2) to estimate their correctness. 
When more than one  reference simplification exists, we calculate the per-transformation F1 scores of the output against each reference, and then keep the highest one as the sentence-level score.
The corpus-level scores are the average of sentence-level scores.

\paragraph{Quality Estimation Features}
Traditional automatic metrics used for SS rely on the existence and quality of references, and are often not enough to analyse the complex process of simplification.
QE leverages both the source sentence and the output simplification to provide additional information on specific behaviours of simplification systems which are not reflected in metrics such as SARI.
EASSE uses QE features from \citet{qa4ts:martin-etal:2018}'s open-source repository\footnote{\url{https://github.com/facebookresearch/text-simplification-evaluation}}.
The QE features currently available are: the compression ratio of the simplification with respect to its source sentence, its Levenshtein similarity, the average number of sentence splits performed by the system, the proportion of exact matches (i.e.\ original sentences left untouched), average proportion of added words, deleted words, and lexical complexity score\footnote{The lexical complexity score of a simplified sentence is computed by taking the log-ranks of each word in the frequency table. 
The ranks are then aggregated by taking their third quartile.}. 

\subsection{Access to Test Datasets}
\label{sec:datasets}
EASSE provides access to three publicly available datasets for automatic SS evaluation (Table~\ref{tab:datasets}): PWKP~\citep{zhu-etal:2010}, TurkCorpus~\citep{xu-etal:2016}, and HSplit~\citep{badbleu:sulem-etal:2018}.
All of them consist of the data from the original datasets, which are sentences extracted from English Wikipedia (EW) articles. 
EASSE can also evaluate system's outputs in other custom datasets provided by the user.

\paragraph{PWKP} \citet{zhu-etal:2010} automatically aligned sentences in 65,133 EW articles to their corresponding versions in Simple EW (SEW). 
Since the latter is aimed at English learners, its articles are expected to contain fewer words and simpler grammar structures than those in their EW counterpart. 
The test set split of PWKP contains 100 sentences, with 1-to-1 and 1-to-N alignments (resp. 93 and 7 instances). 
The latter correspond to instances of sentence splitting.
Since this dataset has only one reference for each original sentence, it is not ideal for calculating automatic metrics that rely on multiple references, such as SARI.

\paragraph{TurkCorpus} \citet{xu-etal:2016} asked crowdworkers to simplify 2,359 original sentences extracted from PWKP to collect multiple simplification references for each one. 
This dataset was then randomly split into tuning (2,000 instances) and test (359 instances) sets.
The test set only contains 1-to-1 alignments, mostly with instances of paraphrasing and deletion.
Each original sentence in TurkCorpus has 8 simplified references. 
As such, it is better suited for computing SARI and multi-reference BLEU scores.

\paragraph{HSplit} \citet{badbleu:sulem-etal:2018} recognised that existing EW-based datasets did not contain sufficient instances of sentence splitting. 
As such, they collected four reference simplifications of this transformation for all 359 original sentences in the TurkCorpus test set. 
Even though SAMSA's computation does not require access to references, this dataset can be used to compute an upperbound on the expected performance of SS systems that model this type of structural simplification.

\begin{table}[tb]
\centering
\resizebox{\columnwidth}{!}{
\begin{tabular}{@{}lrcr@{}}
\toprule
    Test Dataset                 &  Instances     & Alignment Type    & References \\
\midrule
    \multirow{2}{*}{PWKP}   & 93            & 1-to-1            & 1  \\
                            & 7             & 1-to-N            & 1  \\
    TurkCorpus		        & 359           & 1-to-1            & 8 \\
    HSplit	                & 359            & 1-to-N            & 4 \\
\bottomrule
\end{tabular}}
\caption{Test datasets available in EASSE. An instance corresponds to a source sentence with one or more possible references. Each reference can be composed of one or more sentences.}
\label{tab:datasets}
\end{table}

\subsection{HTML Report Generation}
EASSE wraps all the aforementioned analyses in a simple comprehensive HTML report that can be generated with a single command.
This report compares the system output with human reference(s) using simplification metrics and QE features. It also plots the distribution of compression ratios or Levenshtein similarities between sources and simplifications over the test set.
Moreover, the analysis is broken down by source sentence length in order to get insights on how the model handles short source sentence versus longer source sentences, e.g.\ \textit{does the model keep short sentences unmodified more often than long sentences?}
This report further facilitates qualitative analysis of system outputs by displaying source sentences with their respective simplifications.
The modifications performed by the model are highlighted for faster and easier analysis. 
For visualisation, EASSE samples simplification instances to cover different behaviours of the systems. 
Instances that are sampled include simplifications with sentence splitting, simplifications that significantly modify the source sentence, output sentences with a high compression rate, those that display lexical simplifications, among others.
Each of these aspects is illustrated with 10 instances.
An example of the report can be viewed at \url{https://github.com/feralvam/easse/blob/master/demo/report.gif}.


\section{Experiments}
\label{sec:experiments}

We collected publicly available outputs of several SS systems (Sec.~\ref{sec:systems}) to evaluate their performance using the functionalities available in EASSE. 
In particular, we compare them using automatic metrics, and provide some insights on the reasoning behind their results (Sec.~\ref{sec:systems_evaluation}).

\subsection{Sentence Simplification Systems}
\label{sec:systems}
EASSE provides access to various SS system outputs that follow different approaches for the task. 
For instance, we include those that rely on phrase-based statistical MT, either by itself  (e.g.~PBSMT-R~\citep{wubben-etal:2012}), or coupled with semantic analysis, (e.g.~Hybrid~\citep{narayan-gardent:2014}). 
We also include SBSMT-SARI~\citep{xu-etal:2016}, which relies on syntax-based statistical MT; \textsc{Dress-Ls}~\cite{zhang-lapata:2017}, a neural model using the standard encoder-decoder architecture with attention combined with reinforcement learning; and DMASS-DCSS~\citep{dmass-dcss:zhao-etal:2018}, the current state-of-the-art in the TurkCorpus, which is based on the Transformer architecture~\citep{transformer:vaswani-etal:2017}.

\subsection{Comparison and Analysis of Scores}
\label{sec:systems_evaluation}
%

\paragraph{Automatic Metrics}
For illustration purposes, we compare systems' outputs using BLEU and SARI in TurkCorpus (with 8 manual simplification references), and SAMSA in  HSplit. 
For calculating Reference values in Table~\ref{tab:autometrics}, we sample one of the 8 human references for each instance as others have done~\citep{zhang-lapata:2017}.

When reporting SAMSA scores, we only use the first 70 sentences of TurkCorpus that also appear in HSplit.\footnote{At the time of this submission only a subset of 70
sentences had been released from HSplit. However, the full corpus will
soon be available in EASSE.} 
This allows us to compute Reference scores for instances that contain structural simplifications (i.e.\ sentence splits).
We calculate SAMSA scores for each of the four manual simplifications in HSplit, and choose the highest as an upper-bound Reference value.
The results for all three metrics are shown in  Table~\ref{tab:autometrics}. 

\begin{table}[!htb]
\centering \small
\begin{tabular}{@{}lrrr@{}}
\toprule
 	                        & \multicolumn{2}{c}{TurkCorpus}        & HSplit \\
\cmidrule(lr){2-3} \cmidrule(l){4-4}
 	 System                 & SARI     & BLEU      & SAMSA     \\
\midrule
    Reference               & 49.88     & 97.41     & 54.00 \\
\midrule
 	PBSMT-R                 & 38.56     & \bf 81.11 & \bf 47.59\\
 	Hybrid                  & 31.40     & 48.97     & 46.68\\
    SBSMT-SARI              & 39.96     & 73.08     & 41.41\\
 	\textsc{Dress-Ls}	    & 37.27     & 80.12     & 45.94\\
 	DMASS-DCSS              & \bf 40.42 & 73.29     & 35.45\\
\bottomrule
\end{tabular}
\caption{Comparison of systems' performance based on automatic metrics.}
\label{tab:autometrics}
\end{table}

\noindent{DMASS-DCSS is the state-of-the-art in TurkCorpus according to SARI. 
However, it gets the lowest SAMSA score, and the third to last BLEU score. 
PBSMT-R is the best in terms of these two metrics. 
Finally, across all metrics, the Reference stills gets the highest values, with significant differences from the top performing systems.}

\paragraph{Word-level Transformations} In order to better understand the previous results, we use the word-level annotations of text transformations (Table~\ref{tab:oplabels}).
Since SARI was design to evaluate mainly paraphrasing transformations, the fact that SBSMT-SARI is the best at performing replacements and second place in copying explains its high SARI score.
DMASS-DCSS is second best in replacements, while PBSMT-R (which achieved the highest BLEU score) is the best at copying. 
Hybrid is the best at performing deletions, but is the worst at replacements, which SARI mainly measures.
The origin of the TurkCorpus set itself could explain some of these observations. 
According to \citet{xu-etal:2016}, the annotators in TurkCorpus were instructed to mainly produce paraphrases, i.e.\ mostly replacements with virtually no deletions. 
As such, copying words is also a significant transformation, so systems that are good at performing it better mimic the characteristics of the human simplifications in this dataset.

\begin{table}[htb]
\centering \small
\begin{tabular}{@{}lrrrr@{}}
\toprule
	System            & Delete      & Move     & Replace    & Copy\\
\midrule
 	PBSMT-R		      & 34.18       & 2.64      & 23.65     & \bf 93.50\\
 	Hybrid 			  & \bf 49.46   & \bf 7.37  & 1.03      & 70.73\\
    SBSMT-SARI        & 28.42       & 1.26      & \bf 37.21 & 92.89\\
 	\textsc{Dress-Ls} & 40.31       & 1.43      & 12.62     & 86.76\\
 	DMASS-DCSS        & 38.03       & 5.10      & 34.79     & 86.70 \\
\bottomrule
\end{tabular}
\caption{Transformation-based performance of the sentence simplification systems in the TurkCorpus test set.}
\label{tab:oplabels}
\end{table}

\paragraph{Quality Estimation Features}
Table~\ref{tab:qe_table} displays a subset of QE features that reveal other aspects of the simplification systems.
For instance, the scores make it clear that Hybrid compresses the input way more than other systems (compression ratio of 0.57 vs. $\geq$0.78 for the other systems) but almost never adds new words (addition proportion of 0.01). 
This additional information explains the high Delete and low Replace performance of this system in Table~\ref{tab:oplabels}.
\textsc{Dress-Ls} keeps the source sentence unmodified $26\%$ of the time, which does not show in the word-level analysis. 
This confirms that QE features are complementary to automatic metrics and word-level analysis.

\begin{table}[htbp]
\centering
\resizebox{\columnwidth}{!}{
\begin{tabular}{@{}lccccc@{}}
\toprule
System & \makecell{Compression\\ratio} & \makecell{Exact\\matches} & \makecell{Additions\\proportion} & \makecell{Deletion\\proportion} \\
\midrule
PBSMT-R    & 0.95            & 0.1           & 0.1                  & 0.11                \\
Hybrid     & \bf 0.57        & 0.03          & 0.01                 & \bf 0.41                \\
SBSMT-SARI & 0.94            & 0.11          & \bf 0.16                 & 0.13                \\
\textsc{Dress-Ls}   & 0.78            & \bf 0.26          & 0.04                 & 0.26                \\
DMASS-DCSS & 0.89            & 0.05          & 0.15                 & 0.21               \\
\bottomrule
\end{tabular}}
\caption{Quality estimation features, which give additional information on the output of different systems.\label{tab:qe_table}}
\end{table}

\paragraph{Report}

\begin{figure}[tb]
    \centering
    \includegraphics[width=\columnwidth]{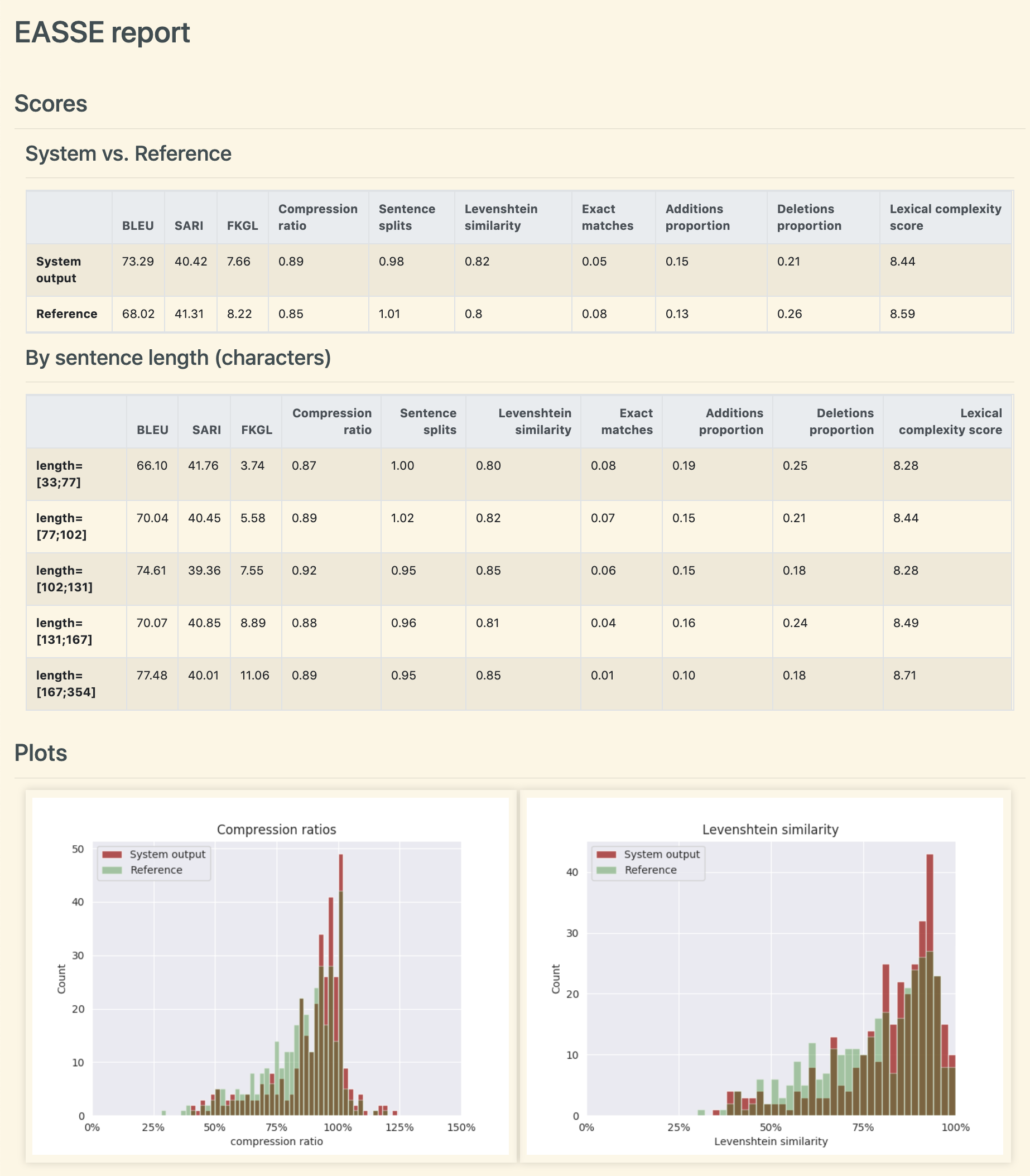}
    \caption{Overview of the HTML report for the DMASS-DCSS system (zoom in for more details).\label{fig:report_screenshot}}
\end{figure}

Figure~\ref{fig:report_screenshot} displays the quantitative part of the HTML report generated for the DMASS-DCSS system.
The report compares the system to a reference human simplification.
The ``System vs. Reference'' table and the two plots indicate that DMASS-DCSS closely matches different aspects of human simplifications,  according to QE features.
This contributes to explaining the high SARI score of the this system in Table~\ref{tab:autometrics}.

\section{Conclusion and Future Work}
EASSE provides easy access to commonly used automatic metrics as well as to more detailed word-level transformation analysis and QE features which allows us to compare the quality of the generated outputs of different SS systems on public test datsets.
We reported some experiments on the use of automatic metrics to obtain overall performance scores, followed by measurements of how effective the SS systems are at executing specific simplification transformations using word-level analysis and QE features.
The former analysis provided  insights about the simplification capabilities of each system, which help better explain the initial automatic scores. 

In the future, we plan to continue developing the transformation-based analysis algorithms, so that more sophisticated transformations could be identified (e.g.\ splitting or subject-verb-object reordering). In addition, we expect to integrate more QE features to cover other aspects of the simplification process (e.g. depth of the dependency parse tree to measure syntactic complexity). 


\bibliography{bibliography}
\bibliographystyle{acl_natbib}

%


\end{document}